\documentclass[lettersize,journal]{IEEEtran}
\pdfoutput=1
\usepackage{amsmath,amsfonts}
\usepackage{algorithmic}
\usepackage{array}
\usepackage[caption=false,font=normalsize,labelfont=sf,textfont=sf]{subfig}
\usepackage{textcomp}
\usepackage{stfloats}
\usepackage{url}
\usepackage[breaklinks=true]{hyperref}  
\usepackage{verbatim}
\usepackage{graphicx}
\usepackage{bm}      
\usepackage{amssymb} 

\newif\ifarXiv
\arXivtrue

\def\BibTeX{{\rm B\kern-.05em{\sc i\kern-.025em b}\kern-.08em
    T\kern-.1667em\lower.7ex\hbox{E}\kern-.125emX}}
\usepackage{balance}
\usepackage{multirow,booktabs,threeparttable,tabularx}  
\usepackage{cite} 
\DeclareSymbolFont{myletters}{OML}{ztmcm}{m}{it}
\DeclareMathSymbol{\lambda}{\mathord}{myletters}{"15}
\newcolumntype{K}[1]{>{\centering\arraybackslash}p{#1}}

\begin{document}
\title{Multiscale Feature Learning Using Co-Tuplet Loss for Offline Handwritten Signature Verification}
\author{Fu-Hsien Huang and Hsin-Min Lu
\thanks{The authors are extremely grateful for the work of Chia-Chun Ku during the early stages of the research. \textit{(Corresponding author: Hsin-Min Lu.)}}
\thanks{Fu-Hsien Huang is with the Department of Information Management, National Taiwan University, Taipei 106, Taiwan (e-mail: d10725004@ntu.edu.tw)}
\thanks{Hsin-Min Lu is with the Department of Information Management, and the Center for Research in Econometric Theory and Applications, National Taiwan University, Taipei 106, Taiwan (e-mail: luim@ntu.edu.tw)}}

\maketitle

\begin{abstract}
Handwritten signature verification, crucial for legal and financial institutions, faces challenges including inter-writer similarity, intra-writer variations, and limited signature samples. To address these, we introduce the MultiScale Signature feature learning Network (MS-SigNet) with the co-tuplet loss, a novel metric learning loss designed for offline handwritten signature verification. MS-SigNet learns both global and regional signature features from multiple spatial scales, enhancing feature discrimination. This approach effectively distinguishes genuine signatures from skilled forgeries by capturing overall strokes and detailed local differences. The co-tuplet loss, focusing on multiple positive and negative examples, overcomes the limitations of typical metric learning losses by addressing inter-writer similarity and intra-writer variations and emphasizing informative examples. The code is available at \url{https://github.com/ashleyfhh/MS-SigNet}. We also present HanSig, a large-scale Chinese signature dataset to support robust system development for this language. The dataset is accessible at \url{https://github.com/hsinmin/HanSig}. Experimental results on four benchmark datasets in different languages demonstrate the promising performance of our method in comparison to state-of-the-art approaches.
\end{abstract}

\section{Introduction}
Handwritten signature verification aims to recognize individuals' signatures for identity verification. This biometric verification approach is commonly accepted by government agencies and financial institutions \cite{RN1}. The handwritten signature verification systems can be classified into two categories based on the modalities of signatures: online and offline signature verification. Online verification involves capturing dynamic characteristics of the signing process, such as velocity and pressure, using specialized devices \cite{RN56}. In contrast, offline verification refers to the static verification of scanned signature images. Since offline verification lacks dynamic characteristics, distinguishing between genuine and forged signatures is inherently more challenging. Moreover, discriminating between genuine signatures and skilled forgeries is difficult due to the high level of imitation similarity (i.e., inter-writer similarity). Practical factors such as significant variations within an individual's signatures (i.e., intra-writer variations or intra-personal variability) and the limited number of available signature samples further complicate the implementation of automatic verification systems \cite{RN2}.

Early offline signature verification systems primarily relied on manual feature extraction methods \cite{RN3, RN4}. To address the challenge of intra-writer variations, capturing regional information from local signature regions has been proposed to provide details for static verification \cite{RN5, RN6, RN7}. However, the traditional process involves sequentially applying independent steps of manual regional feature extraction, region similarity estimation, and similarity verification. These methods overlook the interdependencies between feature extraction and similarity measurement, resulting in suboptimal performance. In recent years, some studies \cite{RN2, RN8, RN9, RN10, RN11, RN12} have proposed the adoption of convolutional neural network (CNN)-based metric learning methods to integrate similarity measurement into automatic feature learning, overcoming the limitations of manual feature extraction. However, existing methods either learn from entire signature images or local regions, failing to exploit the complementary nature of global and regional information. Additionally, most of these methods trained with typical metric learning losses, such as contrastive loss \cite{RN13} and triplet loss \cite{RN14, RN15}, tend to suffer from slow convergence and bad local minima due to the pair or triplet sampling problem \cite{RN16, RN17}.

To overcome the limitations of previous studies, we propose a MultiScale Signature feature learning Network (MS-SigNet) with a new metric learning loss called co-tuplet loss for offline signature verification. MS-SigNet simultaneously considers global and regional information in handwritten signatures, dividing deep feature maps into dual-orientation regions to learn local differences. Learning global representations aims to capture the overall information on signature strokes and configuration. However, given the high similarity between genuine signatures and skilled forgeries, it is also necessary to learn regional representations to explore local details. The proposed MS-SigNet can capture and integrate global and regional information from various spatial scales to generate discriminative features. Additionally, we address the thin and sparse nature of signature strokes in images by employing a multilevel feature fusion (MFF) module to aggregate low-level detailed and high-level semantic information. We also introduce a global-regional channel attention (GRCA) module that guides the network to focus on important information by considering interactions between various spatial scales. The design of our proposed signature verification system is based on the writer-independent (WI) approach. In contrast to the writer-dependent (WD) approach that constructs the signature verification system separately for each writer, WI builds a generic verification system for all writers' signatures. Our WI-based system has the advantages of leveraging information across signatures from different writers and requiring no updates for new writers.

\begin{figure*}[!t]
	\centering
	\includegraphics[width=0.8\textwidth]{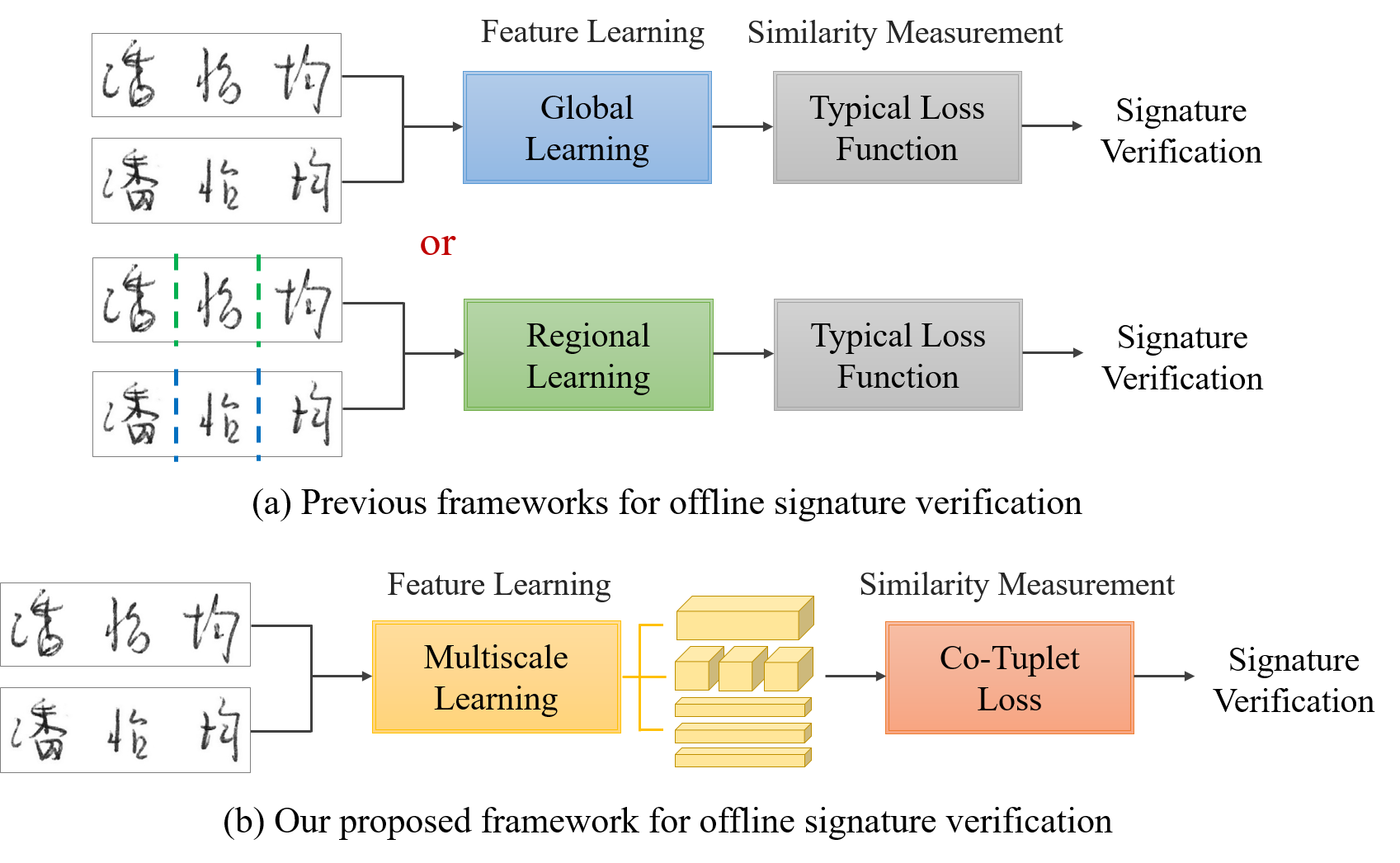}
	\caption{Comparison between previous frameworks and our proposed framework for offline handwritten signature verification}
	\label{fig1}
\end{figure*}

For effective similarity measurement, we propose the co-tuplet loss, a novel metric learning loss, to learn the distance metric for handwritten signature verification. The proposed co-tuplet loss aims to transform input features into a feature space where genuine signatures from the same writer are close to each other while corresponding forgeries are far away from genuine ones. Unlike the typical triplet loss \cite{RN14, RN15}, the proposed co-tuplet loss simultaneously considers multiple genuine signatures and multiple forgeries to learn similarity metrics. This effectively addresses issues related to intra-writer variability and inter-writer similarity. Additionally, we emphasize the importance of batch construction and example selection to focus the training process on informative examples. It is worth noting that our approach combines the feature learning and similarity measurement steps and can be optimized end-to-end. Fig. \ref{fig1} depicts the comparison between previous frameworks and our proposed framework for offline handwritten signature verification.

For training and evaluation, we utilize four offline handwritten signature datasets in different languages. Three of these datasets are publicly available: CEDAR \cite{RN18} with English signatures, BHSig-Bengali \cite{RN19} with Bengali signatures, and BHSig-Hindi \cite{RN19} with Hindi signatures. To address the lack of large-scale public offline Chinese signature datasets, we create the HanSig dataset, consisting of 35,400 signature samples from 238 writers. HanSig surpasses existing public offline Chinese signature datasets in terms of the number of signature samples, facilitating the development of signature verification systems. Experimental results demonstrate the promising performance of our proposed MS-SigNet with co-tuplet loss compared to state-of-the-art methods.

In summary, our contributions can be listed as follows:
\begin{itemize}
	\item{We propose a multiscale feature learning method to generate discriminative features for offline handwritten signature verification. To our best knowledge, this is the first study that adopts deep end-to-end learning to automatically learn and integrate multiple spatial information for offline signature verification.}
	\item{We propose a new metric learning loss that enhances discriminative capability and facilitates better convergence of our network. This loss enables the training process to pay attention to informative examples and effectively tackles challenges associated with intra-writer signing variation and inter-writer similarity, resulting in improved performance.}
	\item{Considering that few large-scale Chinese signature datasets are publicly available, we present the HanSig dataset, a large-scale offline Chinese signature dataset. Such datasets, which consider writers' signing variations, are crucial for developing robust verification systems for this script.}
\end{itemize}

\section{Related Research}
\subsection{Offline handwritten signature verification}\label{research-sub1}
Given the wide use of the offline handwritten signatures, many new approaches for offline signature verification have been developed in the last ten years \cite{RN20}. Most early work \cite{RN3, RN4} relied on manual feature extraction methods to capture signature stroke variations. However, the feature extraction process is easily disturbed by noise, leading to a limited capacity to extract complex features \cite{RN21}. In recent times, there has been a growing interest in utilizing automatic feature extraction methods, particularly CNNs, to learn representations from signature images. These methods have effectively overcome the limitations of manual feature extraction. Several studies \cite{RN22, RN23, RN24} employed CNNs as feature extractors, followed by training separate classifiers for forgery detection. Wei et al. \cite{RN25} introduced a four-stream CNN to focus on the sparse stroke information.

For the improvement of offline signature verification, several studies have concentrated on regional information and local details to capture static properties. To address intra-writer variations, Pirlo and Impedovo \cite{RN5} and Malik et al. \cite{RN6} discovered that stable signature regions exhibit similar patterns among signatures from the same signer. Sharif et al. \cite{RN7} combined global features with local features from 16 image parts. While these methods provided additional information for static signature verification, the separation of manual feature extraction and similarity measurement did not guarantee optimal performance. Liu et al. \cite{RN2} proposed a region-based deep learning network that solely used local regions as inputs to obtain signature features.

In contrast to previous works, we propose an offline signature verification system that automatically learns feature representations from both the entire image and local regions. Unlike Liu et al. \cite{RN2}, our method combines global and regional information and divides deep feature maps into dual-orientation regions. This enables us to aggregate features from multiple scales and improve robustness against misalignment issues. Moreover, our system integrates similarity measurement with feature learning, thereby enhancing the entire training process compared to previous methods \cite{RN5, RN6, RN7, RN22, RN23, RN24}.

\subsection{Metric learning-based methods}\label{research-sub2}
Recently, there has been an increased focus on employing metric learning-based methods to learn similarity and dissimilarity for feature representations. The objective of these methods is to learn a good distance metric that transforms input features into a new feature space, where instances belonging to the same class are close together and those from different classes are far apart \cite{RN26, RN27}. Commonly-used metric learning functions for learning pairwise similarities include contrastive loss \cite{RN13} and triplet loss \cite{RN14, RN15} among others. Deep learning methods that integrate metric learning into feature learning have found wide applications in various domains, including face recognition \cite{RN28-New} and person re-identification \cite{RN27}. 

In the field of offline handwritten signature systems, metric learning-based methods have also shown promising results. Soleimani et al. \cite{RN26} and Rantzsch et al. \cite{RN8} were among the early researchers who introduced metric learning into signature verification. Dey et al. \cite{RN9}, Xing et al. \cite{RN10}, and Liu et al. \cite{RN2} employed the Siamese network \cite{RN29} for metric learning. Some studies have proposed improvements to existing metric learning structures to enhance the robustness of offline signature verification. For instance, Maergner et al. \cite{RN11} combined a triplet loss-based CNN with the graph edit distance approach. Wan and Zou \cite{RN12} and Zhu et al. \cite{RN30} respectively developed a dual triplet loss and a point-to-set (P2S) metric to improve discrimination between genuine signatures and skilled forgeries.

The previous research on handwritten signature verification mainly employed typical metric learning losses or developed improved losses based on similar concepts. However, these losses often suffer from unstable and slow convergence due to inherent sampling problem \cite{RN16, RN17}. To address these limitations of typical metric learning losses, we propose a new metric learning loss. This loss shares similarities with previous tuplet-based losses such as the multi-class N-pair loss \cite{RN16} and the tuplet margin loss \cite{RN31}. However, we introduce a unique example selection and mining strategy specifically tailored for the signature verification task to facilitate better convergence.

\subsection{Main public offline signature datasets}\label{research-sub3}
According to Hameed et al. \cite{RN21}, the CEDAR \cite{RN18}, GPDS \cite{RN32}, and MCYT-75 \cite{RN33} datasets are among the most commonly adopted Western signature datasets. GPDS stands out for its large number of synthetic signatures. In contrast, CEDAR and MCYT-75 include English and Spanish signatures collected from writers, respectively. UTSig \cite{RN35} is a frequently used Persian signature dataset. Additionally, the BHSig260 dataset \cite{RN19} offers two subsets comprising signatures in Bengali and Hindi languages. However, when it comes to offline Chinese handwritten signatures, there is a scarcity of publicly available datasets. Currently, the SigComp2011 \cite{RN34} and ChiSig \cite{RN36} datasets are the only existing public datasets for offline Chinese signatures. SigComp2011 has only 1,177 signature samples. In comparison, ChiSig is a new dataset that contains a more substantial number of samples, with a total of 10,242 signatures.

Considering that the characteristics of handwritten signatures differ across languages and scripts due to their unique writing styles \cite{RN25}, it is impractical to train a Chinese signature verification system using Western datasets. Moreover, the development of a realistic signature verification system requires the consideration of signature variability to avoid overfitting \cite{RN20}. Therefore, we are motivated to create a new offline Chinese signature dataset that contains more samples and incorporates signature variability for each writer.

\section{Proposed Method}\label{method}
In this section, we introduce the MultiScale Signature feature learning Network (MS-SigNet), the handwritten signature verification method proposed in this study. Additionally, we introduce a novel metric learning loss called co-tuplet loss, which aims to improve the discriminative capability of the learned features for signature verification. Finally, we elaborate on the decision-making process in our signature verification system.

\begin{figure*}[!t]
	\centering
	\includegraphics[width=1.0\textwidth]{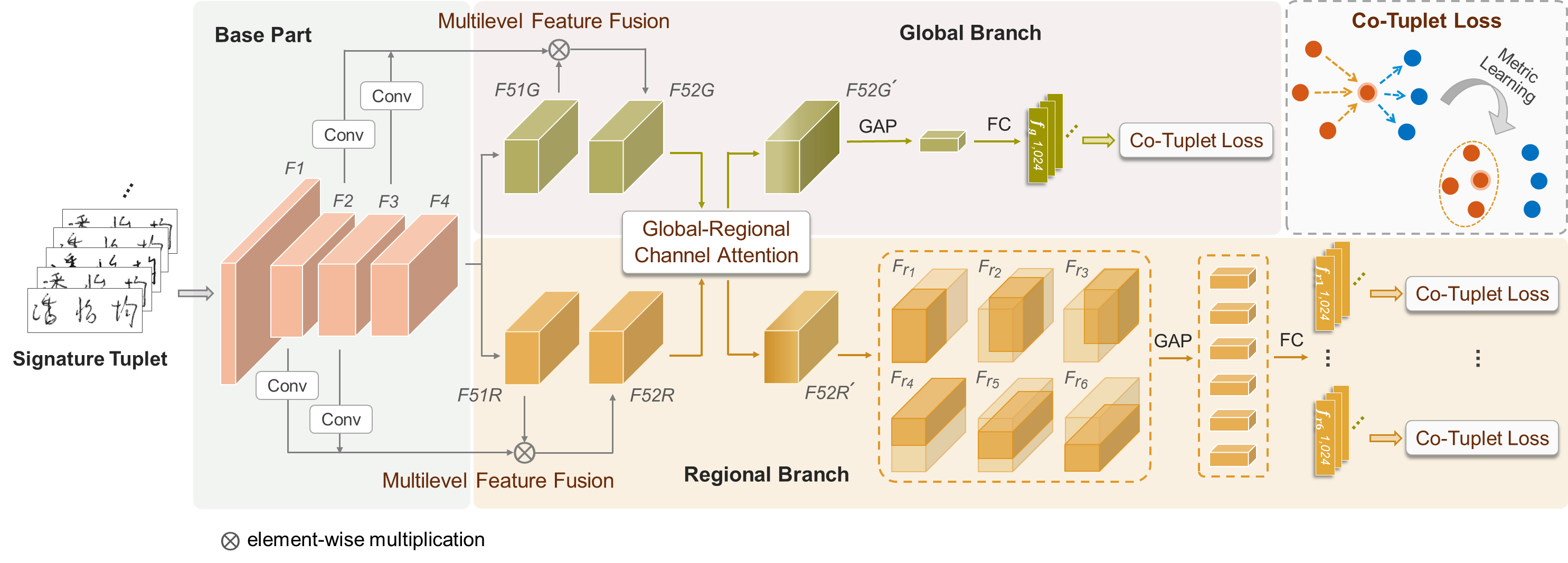}
	\caption{Overall architecture of MS-SigNet. Each set of the corresponding features generated from the proposed network is trained with individual co-tuplet losses}\centering
	\label{fig2}
\end{figure*}

\subsection{Overall architecture}\label{method-sub1}
Fig. \ref{fig2} illustrates the overall architecture of MS-SigNet. To integrate the discriminative signature information of different spatial scales, we propose to automatically learn robust feature representations from both the global and dual-orientation regional branches, which sets it apart from existing methods that rely on manual region feature extraction \cite{RN5, RN6, RN7} and focus solely on local regions \cite{RN2}. We modify the structure of SigNet-F \cite{RN22} as the CNN backbone to build our branches and modules. The layers of the backbone are summarized in Table \ref{tab:1}. We add rectified linear unit (ReLU) activation function and batch normalization (BN) after each convolutional layer to address issues such as vanishing gradients and overfitting. Consider a set of genuine signature images from a specific writer and their corresponding forged counterparts. We refer to this set as a ``signature tuplet'' in the subsequent discussion. Each input image, denoted as $x$, belonging to this signature tuplet passes through the base part to sequentially generate the output feature maps, $F1$, $F2$, $F3$, and $F4$. After the Conv4 layer, the network splits into the global and regional branches and learns to generate feature maps $F51G$ and $F51R$ in the respective branches. By jointly optimizing the global and regional branches, we can enhance the feature learning capability of the base part during the training process. Finally, by training each set of global and regional features with our proposed co-tuplet loss, we obtain the discriminative global and regional embeddings.

\begin{table}
	\caption{Summary of the backbone of MS-SigNet}\label{tab:1}
	\begin{tabular}[h]{lll}
		\toprule
		
		Layer & Size & Stride \& Padding\\
		\midrule
		Input & 1 $\times$ 150 $\times$ 220 &	 \\
		Convolution (Conv1) & 96 $\times$ 11 $\times$ 11 & s = 1, p = 0  \\
		Pooling & 96 $\times$ 3 $\times$ 3 & s = 2  \\
		Convolution (Conv2) & 256 $\times$ 5 $\times$ 5 & s = 1, p = 2 \\
		Pooling	& 256 $\times$ 3 $\times$ 3 & s = 2 \\
		Convolution (Conv3) & 384 $\times$ 3 $\times$ 3 & s = 1, p = 1 \\
		Convolution (Conv4) & 384 $\times$ 3 $\times$ 3 & s = 1, p = 1 \\
		Convolution (Conv5) & 256 $\times$ 3 $\times$ 3 & s = 1, p = 1 \\
		Pooling & 256 $\times$ 3 $\times$ 3 & s = 2 \\
		Global average pooling (GAP) & 256 $\times$ 1 $\times$ 1 & \\ 
		Fully connected (FC) & 1,024 & \\ 
		\bottomrule
	\end{tabular}
	\footnotetext{Except for the fully connected layer, all sizes are described in \textit{depth} $\times$ \textit{height} $\times$ \textit{width}.}
\end{table}

\subsection{Multilevel feature fusion}\label{method-sub2}
We observed that the signature strokes in the images are thin and sparse compared with general object images. In a typical CNN structure, the low-level features generated from the early layers contain more detailed information, while the high-level features generated from the deep layers have rich semantic information. However, information loss of stroke details is inevitable after several convolution and downsampling operators. In order to retain the detailed information of signature strokes, we propose a multilevel feature fusion (MFF) mechanism that combines low-level features with high-level features and then passes the aggregated information to subsequent layers. Here, the feature maps $F2$ and $F3$ in the base part and $F51G$ in the global branch are fused to obtain the feature maps $F52G$. Likewise, $F2$ and $F3$ in the base part and $F51R$ in the regional branch are fused to generate $F52R$. We can express the fusion operations as follows:
\begin{align}
	F52G &= \varepsilon^{3\times3}(F2) \circ \varepsilon^{3\times3}(F3) \circ F51G,\\
	F52R &= \varepsilon^{3\times3}(F2) \circ \varepsilon^{3\times3}(F3) \circ F51R,
\end{align}
\noindent where $\varepsilon^{3\times3}$ is the convolution operation with a kernel size of $3 \times 3$ and a stride of 2 to transfer $F2 \in \mathbb{R}^{C \times H' \times W'}$  and $F3 \in \mathbb{R}^{C' \times H' \times W'}$ to the same shape as $F51G$ and $F51R \in \mathbb{R}^{C \times H \times W}$. The operator $\circ$ denotes the element-wise multiplication.

In contrast to the fusion of only the base part features, we propose a different fusion mechanism that utilizes both the global and regional features for their respective branches. Thus, the detailed information of signature strokes can complement to the high-level features in each branch. In addition, instead of the commonly-used concatenation, we employ a multiplicative operation as the fusion strategy. Multiplication has advantages over concatenation as it allows the gradients of each layer to be correlated with the gradients of the other layers during gradient computation \cite{RN37}. By using multiplication, features at different levels can depend on and interact with each other during the training process.

\subsection{Global-regional channel attention}\label{method-sub3}
In order to extract essential global and regional feature representations, we propose a global-regional channel attention (GRCA) module to guide our model in focusing on specific signature information. GRCA simultaneously learns the attention weights for global and regional features by considering their interactions and relative importance. Drawing inspiration from previous attention mechanisms \cite{RN37, RN38}, we design GRCA tailored for our two-branch structure to facilitate the signature verification task.

\begin{figure}[!t]
	\centering
	\includegraphics[width=0.48\textwidth]{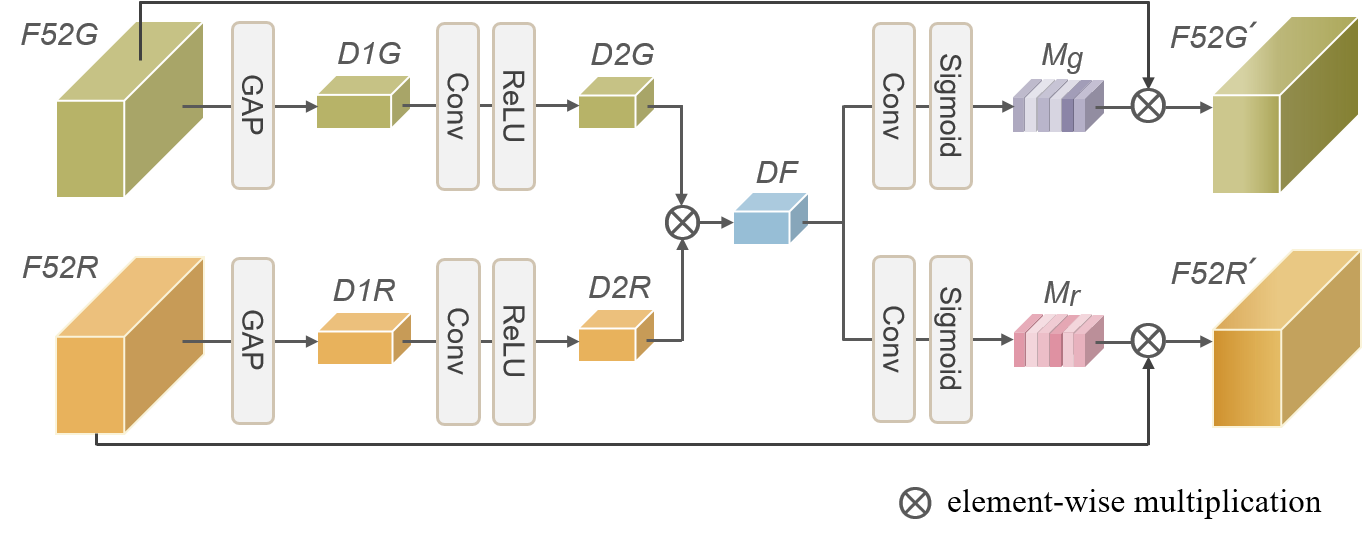}
	\caption{Detailed structure of the GRCA module}
	\label{fig3}
\end{figure}

As shown in Fig. \ref{fig3}, we perform global average pooling (GAP) on the feature maps $F52G \in \mathbb{R}^{C \times H \times W}$ in the global branch and $F52R \in \mathbb{R}^{C \times H \times W}$ in the regional branch to compress the spatial information of each channel into one channel descriptor. We obtain two channel descriptors $D1G \in \mathbb{R}^{C \times 1 \times 1}$ and $D1R \in \mathbb{R}^{C \times 1 \times 1}$:
\begin{align}
	D1G^c &= \phi(\mathcal{Z}^c_g) = \frac{1}{HW}\sum^H_{i=1}\sum^W_{j=1}\mathcal{Z}^c_g(i,j),\\
	D1R^c &= \phi(\mathcal{Z}^c_r) = \frac{1}{HW}\sum^H_{i=1}\sum^W_{j=1}\mathcal{Z}^c_r(i,j),
\end{align}
\noindent where $\phi$ is the GAP operation, $\mathcal{Z}^c_g$ is the $c$-th channel of $F52G$, and $\mathcal{Z}^c_r$ is the $c$-th channel of $F52R$, for $c=1,\ldots,C$. For attention map learning, we first use a convolutional layer with a kernel size of $1 \times 1$ followed by a ReLU activation function to convert the channel descriptors $D1G$ into $D2G \in \mathbb{R}^{V \times 1 \times 1}$ and $D1R$ into $D2R \in \mathbb{R}^{V \times 1 \times 1}$. We set $V<C$, such that the dimension reduction operation can reduce computational and parameter overhead. Subsequently, we combine $D2G$ and $D2R$ to obtain the fused descriptors ${DF} \in \mathbb{R}^{V \times 1 \times 1}$ using the multiplicative operation. The fusion of $D2G$ and $D2R$ enables the simultaneous generation of channel-wise attention for global and regional features based on their relative importance.

Next, we perform a dimension recovery operation for regional and global branches using convolution with a kernel size of $1 \times 1$, and we obtain the normalized global and regional attention maps $M_g$ and $M_r \in \mathbb{R}^{C \times 1 \times 1}$ using the sigmoid function. The entire process of attention map learning can be expressed as follows:
\begin{align}
	M_g &= \sigma(\varepsilon^{1\times1}(\gamma(\varepsilon^{1\times1}(D1G))\circ\gamma(\varepsilon^{1\times1}(D1R)))),\\ 
	M_r &= \sigma(\varepsilon^{1\times1}(\gamma(\varepsilon^{1\times1}(D1G))\circ\gamma(\varepsilon^{1\times1}(D1R)))),
\end{align}
\noindent where $\varepsilon^{1 \times 1}$ is the convolution operation with a kernel size of $1 \times 1$, $\gamma$ represents the ReLU function, and $\sigma$ represents the sigmoid function. We finally multiply each channel of $F52G$ by each weight value of $M_g$ to obtain $F52G' \in \mathbb{R}^{C \times H \times W}$ and perform the same operation on $F52R$ and $M_r$ to obtain $F52R'\in \mathbb{R}^{C \times H \times W}$.

\subsection{Multiscale feature learning}\label{method-sub4}
We propose automatic multiscale feature learning to address the limitations of learning from a single branch. Here, ``multiscale'' refers to multiple scales (i.e., sizes) of global and dual-orientation regional feature maps. By learning complementary feature representations from various spatial scales, our method can effectively handle the challenges posed by high inter-writer similarity and intra-writer variations. To capture signature stroke information from the entire image, we first conduct the GAP operation over the feature maps $F52G' \in \mathbb{R}^{C \times H \times W}$ in the global branch, as shown in Fig. \ref{fig2}. Subsequently, we use a fully connected (FC) layer with the output dimension of 1,024 to generate the global embedding $f_g$ for a signature image. Instead of using multiple FC layers, we adopt GAP followed by a single FC layer to reduce the model parameters and to address overfitting issues. Additionally, we apply L2-Normalization before the output layer to mitigate the impact of scale variability in the data, which enhances training stability and boosts performance. The process of global feature learning can be formulated as follows:
\begin{align}
	\phi(\mathcal{U}^c_g) &= \frac{1}{HW}\sum^H_{i=1}\sum^W_{j=1}\mathcal{U}^c_g(i,j),\\
	f_g &= \eta([\phi(\mathcal{U}^1_g),\phi(\mathcal{U}^2_g),...,\phi(\mathcal{U}^C_g)]),
\end{align}
\noindent where $\phi$ is the GAP operation, $\eta$ is the FC layer, and $\mathcal{U}^c_g$ is the $c$-th channel of $F52G'$, for $c=1,\ldots,C$.

Inspired by Pirlo and Impedovo \cite{RN5}, we propose dual-orientation regional feature learning to complement global feature learning. Rather than using a fixed-sized sliding window along one direction of an input image for region segmentation, we propose to divide the deep feature maps into regions with different scales along horizontal and vertical orientations. This dual-orientation regional feature learning can effectively capture more localized differences and dissimilarities between genuine and skilled-forged signatures. Our method does not require input segmentation and does not increase the number of model inputs, resulting in improved efficiency, particularly for larger datasets.

We first divide the feature maps $F52R' \in \mathbb{R}^{C \times H \times W}$ into three overlapping vertical regions $F_{r_1}$, $F_{r_2}$, and $F_{r_3} \in \mathbb{R}^{C \times H \times W''}$ $(W''<W)$ from left to right, as shown in the regional branch of Fig. \ref{fig2}. We also divide the feature maps $F52R'$ into three overlapping horizontal regions $F_{r_4}$, $F_{r_5}$, and $F_{r_6} \in \mathbb{R}^{C \times H'' \times W}$ $(H''<H)$ from top to bottom. To address the potential misalignment, we make adjacent regions overlap each other. The process of region division can be expressed as follows:
\begin{align}
	F_{r_m} &= \psi_H(F52R',m),{\quad} m \in \{1,2,3\},\\
	F_{r_n} &= \psi_V(F52R',n),{\quad} n \in \{4,5,6\},
\end{align}
\noindent where $\psi_H$ denotes the feature map division operation in the horizontal orientation, and $\psi_V$ is the same operation in the vertical orientation. Specifically, we empirically set $W''=13$ and the overlap between $F_{r_m}$ to be 7 pixels in width, and we empirically set $H''=8$ and set the overlap between $F_{r_n}$ to be 4 pixels in height.

Finally, we obtain six regions with varying scales and conduct GAP operations over each of them, followed by an FC layer to generate 1024-dimensional regional embeddings, $f_{r_m}$, $m \in \{1,2,3\}$, and $f_{r_n}$, $n \in \{4,5,6\}$, for each signature image. The process of generating regional embeddings can be formulated as follows:
\begin{align}
	\phi(\mathcal{U}^c_{r_m}) &= \frac{1}{HW}\sum^H_{i=1}\sum^{13}_{j=1}\mathcal{U}^c_{r_m}(i,j),\\
	f_{r_m} &= \eta([\phi(\mathcal{U}^1_{r_m}),\phi(\mathcal{U}^2_{r_m}),...,\phi(\mathcal{U}^C_{r_m})]),\\
	\phi(\mathcal{U}^c_{r_n}) &= \frac{1}{HW}\sum^8_{i=1}\sum^W_{j=1}\mathcal{U}^c_{r_n}(i,j),\\
	f_{r_n} &= \eta([\phi(\mathcal{U}^1_{r_n}),\phi(\mathcal{U}^2_{r_n}),...,\phi(\mathcal{U}^C_{r_n})]),
\end{align}
\noindent where $\phi$ is the GAP operation, $\eta$ is the FC layer, $\mathcal{U}^c_{r_m}$ is the $c$-th channel of $F_{r_m}$, and $\mathcal{U}^c_{r_n}$ is the $c$-th channel of $F_{r_n}$, for $c=1,\ldots,C$. Similar to the global embedding, we apply L2-Normalization before the output layer for generating regional embeddings. We demonstrate the performance improvement achieved through multiscale feature learning and visualize the differences in their effects in Section \ref{experiments}.

\subsection{Co-tuplet loss}\label{method-sub5}
\subsubsection{Limitations of typical loss functions}\label{method-subsub1}
Previous automatic signature verification methods \cite{RN9, RN11, RN12, RN22, RN23, RN24, RN39, RN40} have employed the classification loss and the typical metric learning loss to learn similarity measurement. However, the methods trained with the classification loss, such as the categorical cross-entropy loss, can only differentiate between genuine signatures of different writers, limiting their application to random forgery detection or preliminary feature extraction for a WD classifier. In contrast, typical metric learning loss functions, such as contrastive loss \cite{RN13} and triplet loss \cite{RN14, RN15}, use a single randomly selected negative example in each update, often resulting in unstable and slow convergence \cite{RN16, RN17}. To address this problem, the tuplet-based loss functions \cite{RN16, RN31} consider multiple negative examples from different classes and aim to increase the inter-class distance by pushing them apart in each update.

In contrast to general object recognition tasks focusing on distinguishing between different classes, handwritten signature verification requires discriminating between positive examples (i.e., genuine signatures) and their corresponding negative examples (i.e., skilled forgeries) within each writer. The challenge lies in the presence of potentially high intra-writer variability in genuine signatures and high inter-writer similarity between genuine and forged signatures. This implies that the distances between most genuine-genuine pairs may be large, while the distances between genuine-forged pairs tend to be small. Using only one positive example in each update is insufficient for considering the distances of multiple positive examples, thus failing to address the intra-writer distance variation effectively. Similarly, relying on a single negative example in each update does not adequately account for the inter-writer similarity between a positive example and the remaining negative examples.

To address the challenges in handwritten signature verification, we propose a new tuplet-based metric learning loss called the co-tuplet loss. It combines the property of tuplet-based loss functions, allowing for the inclusion of multiple negative examples to learn an appropriate distance metric between genuine and forged signatures. Additionally, it employs multiple positive examples, enabling the model to learn the relationship between genuine signatures within each writer, which was previously missing in tuplet-based loss functions \cite{RN16, RN31}.

\subsubsection{Proposed loss function}\label{method-subsub2}
We define a tuplet as $\{x_a,x_{p_1},x_{p_2},\ldots,x_{p_k},x_{n_1},x_{n_2},\ldots,x_{n_k}\}$, where $x_a$, $x_p$, and $x_n$ refer to the anchor, positive, and negative examples, respectively. The integer $k$ is the number of positive and negative examples in a mini-batch. In this study, the anchor and positive examples are genuine signatures signed by a writer, while the negative examples are corresponding skilled forgeries. We aim to shorten the intra-writer distance and enlarge the inter-writer distance in the embedding space. To achieve this, we jointly consider the distances of multiple positive and negative examples from the same anchor. The formulation of the co-tuplet loss is as follows:
\begin{equation}
	\label{deqn_ct}
	\mathcal{L}_{ct} = \log[1+\sum_{i\in\mathcal{S}(\mathcal{P})}\exp(d^+_i-d^-_h)+\sum_{j\in\mathcal{S}(\mathcal{N})}\exp(d^+_h-d^-_j)],
\end{equation}
\noindent where $\mathcal{S}(\mathcal{P})$ and $\mathcal{S}(\mathcal{N})$ are the sets of positive and negative example indices for which the positive and negative examples satisfy our mining strategy described in the next subsection, $d$ indicates the squared Euclidean distance used as the distance metric; and $d^+_i$, $d^-_j$, $d^+_h$, and $d^-_h$ are defined as follows:
\begin{align}
	d^+_i &= \|f(x_a)-f(x_{p_i})\|^2_2,\\
	d^-_j &= \|f(x_a)-f(x_{n_j})\|^2_2,\\
	d^+_h &= \max_{\ell=1 \ldots k} \|f(x_a)-f(x_{p_{\ell}})\|^2_2,\\
	d^-_h &= \min_{\ell=1 \ldots k} \|f(x_a)-f(x_{n_{\ell}})\|^2_2,
\end{align}
\noindent where $f(\cdot)$ represents the feature embedding of an input example. Among the positive examples in a mini-batch, $d^+_h$ is the distance between the anchor and the hardest positive example. Similarly, among the negative examples in a mini-batch, $d^-_h$ is the distance between the anchor and the hardest negative example.

\begin{figure*}[!t]
	\centering
	\includegraphics[width=0.85\textwidth]{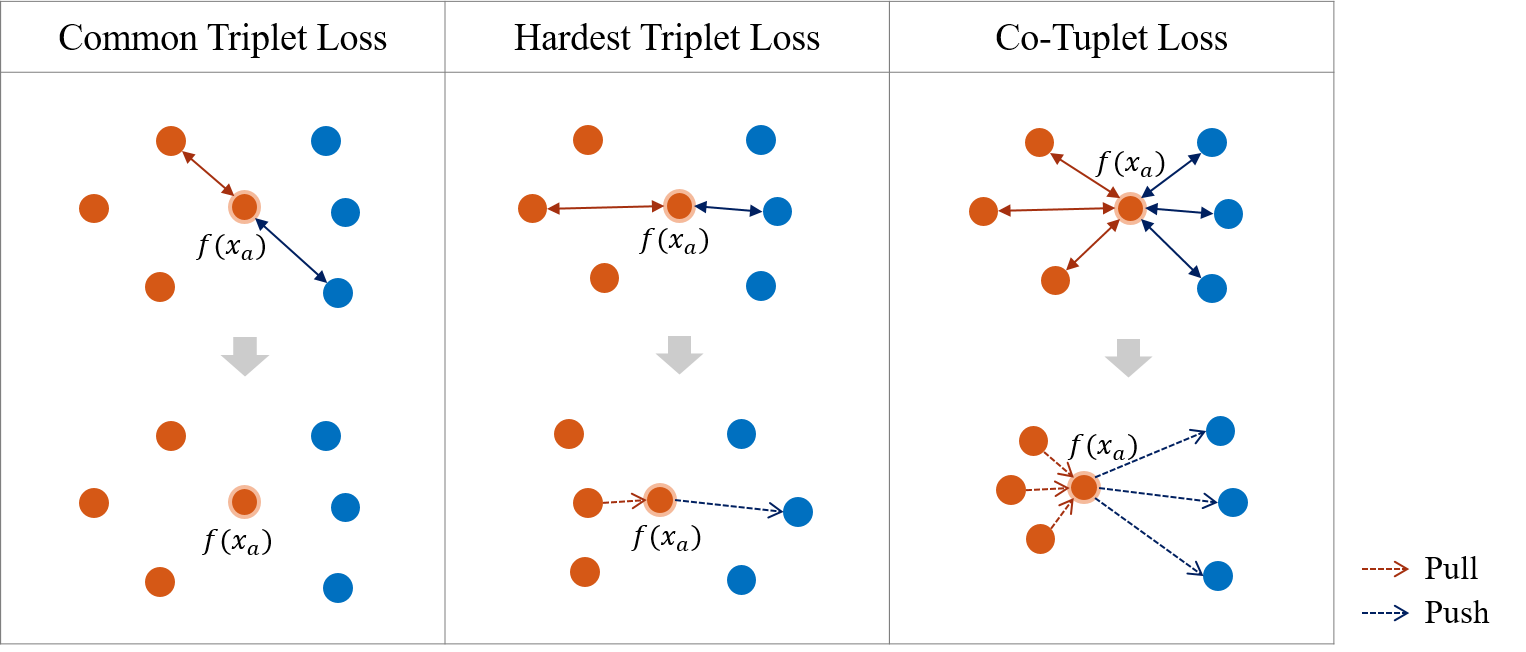}
	\caption{Example of the distance learning process of the common triplet loss, hardest triplet loss, and co-tuplet loss}
	\label{fig4}
\end{figure*}

The proposed loss comprises two parts, pulling and pushing parts. We design the pulling part to decrease the intra-writer distance by pulling the selected positive examples closer to the anchor using $d^-_h$ as the reference distance. In addition, we use the pushing part to increase the inter-writer distance by pushing the selected negative examples farther from the anchor using $d^+_h$ as the reference distance. This reduces the computational complexity from quadratic to linear compared to pairwise comparisons between positive and negative examples. Given that the proposed tuplet-based loss relies on the co-existence of multiple positive and negative examples to learn distance metrics, we refer to it as the ``co-tuplet loss.''

Fig. \ref{fig4} provides a visual representation of the differences in the distance learning process between the triplet loss, the hardest triplet loss, and the co-tuplet loss. The triplet loss only partially utilizes the distance information of batch examples, as many easy triplets that already satisfy the triplet constraint do not contribute to the training process. Similarly, the hardest triplet loss, which focuses on the hardest positive and negative examples, neglects optimizing the distances from the anchor to the remaining positive and negative examples. In contrast, the proposed co-tuplet loss simultaneously considers multiple positive and negative examples in a mini-batch for distance optimization. This allows the signature verification system to learn to pull genuine signatures belonging to the same writer close together, while pushing forgeries far away from them.

\subsubsection{Batch construction and constraint mining strategy}\label{method-subsub3}
To construct a training mini-batch, we first randomly select \textit{w} genuine signatures without replacement as anchor examples. For each anchor, $k$ genuine signatures are randomly sampled (excluding the anchor itself) as positive examples from the same writer. Similarly, $k$ negative examples are randomly sampled from the corresponding forgeries. Together, the anchor, positive examples, and negative examples form a signature tuplet. We repeat this process to generate \textit{w} tuplets for the training mini-batch.

To select training examples, we propose a constraint mining strategy that focuses the learning process on informative signature examples rather than trivial ones. To the best of our knowledge, this mining strategy is not considered in existing tuplet-based losses \cite{RN16, RN31}. We identify that very easy positive and negative examples still contribute to the loss values in Eq. \eqref{deqn_ct}, even though they provide uninformative and redundant information for embedding learning. Hence, we employ the constraint mining strategy to select informative examples. The selected positive and negative examples must satisfy the following constraints:
\begin{align}
	d^+_i \geq d^-_h-\delta,\\
	d^-_j \leq d^+_h+\delta,
\end{align}
\noindent where $\delta>0$ is a constraint margin. Here an appropriate selection of $\delta$ ensures that the optimization process does not concentrate on useless information and facilitates our model in learning the accurate mapping of intra-writer and inter-writer distances.

\subsubsection{Gradient computation}\label{method-subsub4}
We can obtain the gradient of the proposed co-tuplet loss $\mathcal{L}_{ct}$ with respect to the model parameters $\bm{\theta}$:
\begin{align}
&\frac{\partial \mathcal{L}_{ct}}{\partial \bm{\theta}} \nonumber \\ 
&= \frac{1}{q}\Big[\sum_{i\in\mathcal{S}(\mathcal{P})}\exp(d^+_i-d^-_h)\frac{\partial(d^+_i-d^-_h)}{\partial \bm{\theta}} \nonumber \\ 
&\quad + \sum_{j\in\mathcal{S}(\mathcal{N})}\exp(d^+_h-d^-_j)\frac{\partial(d^+_h-d^-_j)}{\partial \bm{\theta}}\Big] \nonumber \\
&\equiv \frac{1}{q}\Big[\sum_{i\in\mathcal{S}(\mathcal{P})} w_{1,i}\frac{\partial(d^+_i-d^-_h)}{\partial \bm{\theta}} + \sum_{j\in\mathcal{S}(\mathcal{N})} w_{2,j}\frac{\partial(d^+_h-d^-_j)}{\partial \bm{\theta}}\Big],
\end{align}
\noindent where
\begin{equation}
	q = 1 + \sum_{i\in\mathcal{S}(\mathcal{P})}\exp(d^+_i-d^-_h) + \sum_{j\in\mathcal{S}(\mathcal{N})}\exp(d^+_h-d^-_j),
\end{equation}
\noindent and $w_{1,i}$ and $w_{2,j}$ denote the weights of the pulling and pushing parts, respectively. As observed from the above gradient, both $w_{1,i}$ and $w_{2,j}$ are exponentially up-weighted with hard examples and exponentially down-weighted with easy ones.

In comparison to typical metric learning losses, our co-tuplet loss offers the advantage of emphasizing informative examples over uninformative ones by assigning unequal weights to examples based on the distance difference. Since skilled-forged signatures often closely resemble genuine signatures for each writer, this weighting scheme promotes the learning of more discriminative features for handwritten signature verification. Furthermore, the co-tuplet loss takes into account moderate examples (i.e., the examples of normal cases) for distance learning. It avoids overweighting extremely hard examples, leading to a more stable optimization process.

\subsection{Signature verification decision}\label{method-sub6}
In our approach, we use individual co-tuplet losses to train each set of the corresponding features instead of a single loss for the concatenation of generated multiscale features. This training strategy enables the model to learn specific information from each part of the signature strokes. For joint multiscale feature learning, we define the overall objective function as follows:
\begin{equation}
	\mathcal{L}_{ct,T} = \mathcal{L}_{ct,g} + \lambda\sum^6_{i=1} \mathcal{L}_{ct,r_i},
\end{equation}
\noindent where $\lambda$ is a hyperparameter used to control the weight of the regional losses.

In the verification stage, namely the test stage, we integrate various spatial information by concatenating the global and regional embeddings into the final embedding for each input image. To make the verification decision, we use a distance threshold $d_{thr}$ to decide whether a given signature pair $\{x_i,x_j\}$ is positive or negative. A positive verification decision indicates that the questioned signature $x_j$ is accepted as genuine with respect to the reference signature $x_i$. Conversely, $x_j$ with a negative decision is regarded as forged with respect to $x_i$. We define the final signature verification decision as follows:
\begin{equation}
	{\text{Decision}} =
	\begin{cases}
		\text{positive},&{\text{if}}\ d(x_i,x_j) \leq d_{thr}\\
		{\text{negative},}&{\text{if}}\ d(x_i,x_j) > d_{thr},
	\end{cases}
\end{equation}
\noindent where $d(x_i,x_j)$ is the squared Euclidean distance between the final embeddings of $x_i$ and $x_j$.

\section{Experiments}\label{experiments}
In this section, we first describe the benchmark datasets. Following that, we elaborate on the data preprocessing, implementation particulars, and the evaluation metrics used for signature verification. Lastly, we present the experimental results and compare them with various state-of-the-art methods.

\subsection{Datasets}\label{exp-sub1}
In order to train and evaluate the effectiveness of our signature verification system, we utilize four offline handwritten signature datasets in different languages. CEDAR \cite{RN18} is an English signature dataset. BHSig260 \cite{RN18} comprises two sub-datasets, one in Bengali and the other in Hindi. The three public datasets are commonly adopted in this field, enabling comparisons with existing methods. Additionally, to demonstrate the applicability of our method to Chinese signatures, we conduct experiments on the newly created HanSig dataset. Table~\ref{tab:2} provides the details of the benchmark datasets. Because the discrimination between genuine and skilled-forged signatures is more challenging compared to detecting random and simple forgeries, we focus on experiments related to detecting skilled forgeries available in these datasets.

\begin{table}[h]
	\caption{Details of the used benchmark datasets}\label{tab:2}%
	\begin{tabular}{@{}lccc@{}}
		\toprule
		Dataset & Language  & \# Writers & \# Signatures \\
		\midrule
		CEDAR \cite{RN18} & English & 55 & 2,640  \\
		BHSig-Bengali \cite{RN19} & Bengali & 100 & 5,400  \\
		BHSig-Hindi \cite{RN19} & Hindi & 160 & 8,640 \\
		HanSig & Chinese & 238 & 35,400 \\
		\bottomrule
	\end{tabular}
\end{table}

\subsubsection{Offline Chinese signature dataset construction}\label{exp-subsub1}
We construct HanSig, a new Chinese signature dataset, to facilitate the development of signature verification systems. We first collected 554,723 names from the public admission lists of college entrance exams between 1996 and 2002. Next, we split the collected names into first and last names and filtered out the names that appeared less than two times. We generated 885 candidate names based on the frequency distributions of the real world's first and last name distribution. Through the process of generating candidate names, we took precautions to avoid potential legal concerns related to personal information. We collected signatures for these candidate names from 238 writers. The writers were asked to practice sufficiently to become proficient before signing. To introduce more variation in the genuine signatures, each name was signed 20 times in three different styles: neat, normal, and stylish. To create skilled forgeries, we requested another writer, acting as the forger, to thoroughly practice and skillfully imitate one writer's genuine signatures. 

\begin{figure}[h]
	\centering
	\includegraphics[width=0.3\textwidth]{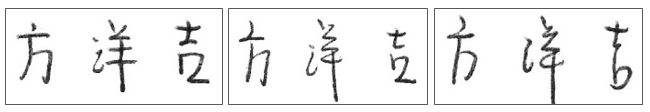}
	\centerline{(a) Examples of collected signatures in three styles}
	\\[5pt]
	\vfill
	\includegraphics[width=0.5\textwidth]{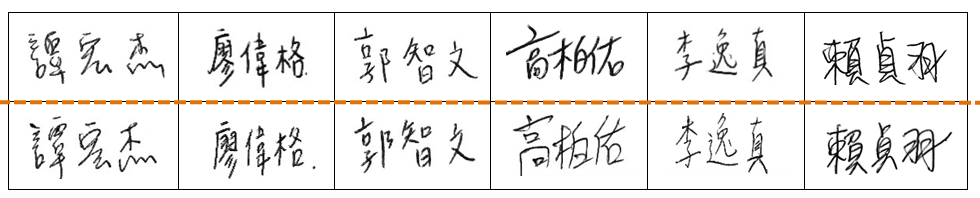}
	\centerline{(b) Examples of collected genuine and forged signatures}
	\\[5pt]
	\caption{Examples of signature images in HanSig. The left, middle, and right images of (a) are the collected signatures written in neat, normal, and stylish styles, respectively. The first row of (b) shows the collected genuine signatures, and the second row of (b) shows the corresponding forged signatures}
	\label{fig5}
\end{figure}

We scanned the collected data using a flatbed scanner to obtain grayscale JPEG files with a resolution of 600 DPI, which is the average resolution adopted in most offline signature datasets. We then cropped the handwritten signature patches from the scanned files. Overall, HanSig consists of a total of 17,700 genuine signatures and an equal number of skilled forgeries. Fig. \ref{fig5} provides examples of the collected signatures in different styles and the genuine and forged signatures.

The HanSig dataset has several valuable characteristics: (1) The generation of names for signatures addresses concerns regarding personal information and privacy while still preserving the distributional characteristics of real names. (2) HanSig incorporates the real-world property of intra-writer variability by including multiple signature styles. (3) HanSig surpasses existing public Chinese signature datasets in terms of the number of signature samples, providing robust training for signature verification systems. (4) HanSig is advantageous for both random and skilled forgery verification tasks.

\subsubsection{Experimental protocol}\label{exp-subsub2}
The CEDAR dataset \cite{RN18} consists of English signatures from 55 writers. Each writer contributed 24 genuine signatures for a specific name and had 24 skilled forgeries generated by forgers. To compare the proposed method with state-of-the-art methods, we follow most previous studies \cite{RN2, RN10, RN25, RN41, RN42} to randomly divide the writers into a training set that includes signatures from 50 writers, and a test set that contains signatures from five writers. We further randomly reserve five writers from the training set for validation. For each writer in the test set, we use one genuine signature as the reference signature and another genuine signature as the questioned signature to form 276 $(24 \times 23/2)$ positive pairs. Additionally, we form 576 $(24 \times 24)$ negative pairs by using one genuine signature as the reference and one forged signature as the questioned signature. Following the strategy of most previous studies, we randomly select the same number of negative pairs as positive pairs to avoid potential bias in performance metrics. The final test data consists of 2,760 signature pairs.

The BHSig-Bengali dataset \cite{RN19} comprises Bengali signatures from 100 writers. Each writer contributed 24 genuine signatures for a specific name, along with 30 skilled forgeries. Following the data splitting scheme in \cite{RN9, RN25, RN52}, we randomly select signatures from 50 writers to form the training set, while the remaining writers' signatures are used for the test set. We further reserve five writers' signatures from the training set for validation. Similarly, for each writer in the test set, we follow the strategy of previous studies to form 276 positive pairs and 276 negative pairs. The final test data comprises 27,600 signature pairs.

The BHSig-Hindi dataset \cite{RN19} contains Hindi signatures from 160 writers. Each writer contributed 24 genuine signatures for a specific name, accompanied by 30 skilled forgeries. Consistent with most previous studies \cite{RN9, RN25, RN43, RN52}, we randomly select signatures from 100 writers to form the training set, while the remaining writers' signatures are used for the test set. Within the training set, we randomly select signatures from five writers to form the validation set. We follow a similar procedure to generate 276 positive test pairs and an equal number of negative test pairs for each writer. The final test data comprises 33,120 signature pairs.

We randomly split HanSig into a training set and a test set. The training set comprises 795 names signed by 213 writers, while the test set includes 90 names signed by 25 writers. From the training set, 20 writers' signatures (78 names) are randomly selected for validation. For each name in the test set, we employ a procedure similar to that used in the three public datasets to create 190 positive pairs and 190 negative pairs. The final test data of HanSig consists of 34,200 signature pairs.

\subsection{Data preprocessing}\label{exp-sub2}
\begin{figure}[h]
	\centering
	\includegraphics[width=0.3\textwidth]{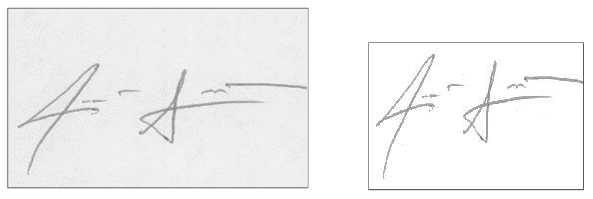}
	\centerline{\footnotesize(a) Genuine signatures}
	\\[5pt]
	\vfill
	\includegraphics[width=0.3\textwidth]{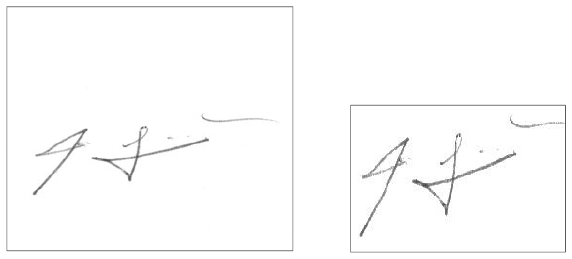}
	\centerline{\footnotesize(b) Forged signature}
	\\[5pt]
	\caption{Examples of CEDAR before and after preprocessing. The left images of (a) and (b) are original signature images, and the right images are preprocessed results}
	\label{fig6}
\end{figure}

To mitigate the influence of background and position variations in the signature images, we perform several data preprocessing steps without deforming the structure of the signatures. Firstly, we convert the signature images to grayscale. Next, we apply Otsu's algorithm \cite{RN44} to transform all background pixel values into 255 while keeping the signature pixels unchanged. This step removes noise in image backgrounds and is critical to datasets (e.g., CEDAR) that have distinct backgrounds in genuine and forged signature images. We also center-crop the signature images and remove the excess blanks around the signatures to eliminate potential misalignment issues caused by signature position variations. The images are resized to the input size of the network using bilinear interpolation. Fig. \ref{fig6} shows significant differences in the samples after these preprocessing steps, particularly in the backgrounds. Finally, the pixel values of the signature images are normalized to a range between 0 and 1.

\subsection{Implementation details and evaluation metrics}
In our model, we experimentally set $C = 256$, $H = 16$, $W = 25$, and $V = 32$. We construct a training mini-batch using \textit{w} $= 18$ for CEDAR, HanSig, and BHSig-Bengali and \textit{w} $= 20$ for BHSig-Hindi. Furthermore, we set $k = 5$ for all datasets. The constraint margin $\delta$ for the constraint mining strategy is empirically set as 0.2 for CEDAR and 0.3 for HanSig, BHSig-Bengali, and BHSig-Hindi. In our experiments, we empirically set $\lambda = 1$ in the objective function as the weight of the regional losses. We apply Adam optimizer with an initial learning rate of 0.001, and the learning rate decays by a factor of 0.5 every 15 epochs. We train our models for a maximum of 90 epochs and use the validation set for early stopping. All our experiments are implemented using the PyTorch framework.

We report the performance using the common evaluation metrics of signature verification: False Reject Rate (FRR), False Accept Rate (FAR), Equal Error Rate (EER), and Area Under the Curve (AUC). FRR refers to the proportion of genuine signatures mistakenly rejected as forgeries. FAR is the proportion of forgeries mistakenly accepted as genuine signatures. EER is the error rate when FRR is equal to FAR, with the adjustment of the decision threshold $d_{thr}$.

\subsection{Method evaluation and analysis}
To assess the performance of the proposed MS-SigNet combined with the co-tuplet loss, we conduct experiments comparing it to a simple baseline model and alternative combinations of losses and models. The evaluation aims to determine the effectiveness of the co-tuplet loss compared to the triplet loss \cite{RN14, RN15} and to understand the relative performance of the MS-SigNet. As a simple baseline, we adopt a VGG-16 \cite{RN45} pretrained on the ImageNet dataset as the feature extractor. Additionally, we train the MS-SigNet with the triplet loss and evaluate its performance for the comparison with the co-tuplet loss. Furthermore, alternative combinations include training the VGG-16 with either the co-tuplet loss or the triplet loss under the same experimental settings.

\begin{table*}[!t]\centering
	\caption{Performance comparison between different combinations of models and losses (evaluation metrics in \%)}\label{tab:3}
	\begin{tabular}[t]{llK{0.8cm}K{0.8cm}K{0.8cm}K{0.8cm}}
		\toprule
		Dataset & Method & FRR & FAR & EER & AUC\\
		\midrule
		\multirow{5}*{CEDAR} & Simple Baseline (Pretrained VGG-16) & 16.67 & 27.75 & 23.08 & 84.82\\
		& VGG-16 with triplet loss & 17.32 & 18.48 & 17.93 & 87.62\\
		& VGG-16 with co-tuplet loss & 17.90 & 15.00 & 16.45 & 90.15\\
		& MS-SigNet with triplet loss & 7.75 & 5.94 & 6.92 & 98.10\\
		& MS-SigNet with co-tuplet loss & {\textbf{3.55}} & {\textbf{3.33}} & {\textbf{3.51}} & {\textbf{99.47}}\\
		\midrule
		\multirow{5}*{BHSig-Bengali} & Simple Baseline (Pretrained VGG-16) & 11.71 & 22.23 & 17.04 & 91.07\\
		& VGG-16 with triplet loss & 12.70 & 12.63 & 12.69 & 94.80\\
		& VGG-16 with co-tuplet loss & 14.44 & 9.08 & 11.96 & 95.75\\ 
		& MS-SigNet with triplet loss & 9.41 & {\textbf{5.21}} & 7.54 & 98.03\\ 
		& MS-SigNet with co-tuplet loss & {\textbf{6.20}} & 5.93 & {\textbf{6.12}} & {\textbf{98.64}}\\
		\midrule
		\multirow{5}*{BHSig-Hindi} & Simple Baseline (Pretrained VGG-16) & 17.28 & 17.71 & 17.52 & 90.64\\   
		& VGG-16 with triplet loss & 14.05 & 15.63 & 14.94 & 92.74\\
		& VGG-16 with co-tuplet loss & 11.91 & 15.94 & 14.04 & 93.86\\
		& MS-SigNet with triplet loss & 9.16 & 8.58 & 8.9 & 96.94\\
		& MS-SigNet with co-tuplet loss & {\textbf{6.56}} & {\textbf{6.76}} & {\textbf{6.68}} & {\textbf{98.28}}\\
		\midrule
		\multirow{5}*{HanSig} & Simple Baseline (Pretrained VGG-16) & 32.43 & 19.66 & 26.31 & 80.94\\ 
		& VGG-16 with triplet loss & 15.60 & 22.40 & 19.07 & 89.47\\
		& VGG-16 with co-tuplet loss & 14.20 & 16.21 & 15.26 & 92.60\\ 
		& MS-SigNet with triplet loss & 9.99 & {\textbf{10.82}} & 10.44 & 95.92\\
		& MS-SigNet with co-tuplet loss & {\textbf{7.69}} & 11.85 & {\textbf{9.93}} & {\textbf{96.38}}\\
		\bottomrule
	\end{tabular}
	\footnotetext{The best results are marked in boldface.}
\end{table*}

Table~\ref{tab:3} reports the experimental results. The MS-SigNet trained with the co-tuplet loss achieves the best performance across all datasets. Compared to the simple baseline, it improves the EER by up to 19.57 percentage points and improves the AUC by up to 15.44 percentage points. Replacing the co-tuplet loss with the triplet loss for training the MS-SigNet results in worsened EER and AUC. Similarly, training the VGG-16 with the co-tuplet loss yields better performance in terms of EER and AUC compared to training it with the triplet loss. Furthermore, comparing the MS-SigNet to the VGG-16 trained with the same loss function, the MS-SigNet consistently outperforms the VGG-16 counterparts. When trained with the co-tuplet loss and the triplet loss, the MS-SigNet exhibits EER improvements ranging from 5.33 to 12.94 and from 5.15 to 11.01 percentage points over the VGG-16, respectively. Overall, the results validate the effectiveness of the MS-SigNet coupled with the co-tuplet loss across different datasets.

\subsection{Ablation studies}

\bigskip\noindent
\textbf{Influence of each module/branch} In this subsection, we first evaluate the importance of each module and branch within our network. To assess the contribution of each component in the proposed MS-SigNet, we systematically remove individual modules and branches from the original framework and analyze the resulting impact on performance. The results of the ablation experiment are reported in Table~\ref{tab:4}. The findings indicate that removing the MFF module results in performance degradation across all datasets, especially with a reduction of 3.23 percentage points for CEDAR. This suggests that our proposed fusion mechanism effectively mitigates information loss caused by layer transmission and retains detailed signature stroke information for both global and regional branches. Additionally, the GRCA module demonstrates a performance improvement ranging from 0.28 to 1.27 percentage points. This highlights the effectiveness of our attention mechanism in focusing on important channel information. The results in Table~\ref{tab:4} also indicate that both the global and regional branches contribute to enhancing signature verification performance. The global branch provides more performance gains for CEDAR, while the regional branch makes a greater contribution to the other datasets. These results suggest that multiscale feature learning are crucial for exploiting their respective advantages.

\bigskip\noindent
\textbf{Validity of selected operations} We also conducted an ablation experiment to evaluate the validity of selected operations in our approach. We compare the performance of the datasets under different operations while maintaining the same experimental settings. Regarding the feature fusion strategy in MFF, the original ``multiplication'' is compared with the conventional ``concatenation''. The results in Table~\ref{tab:5} demonstrate that using ``multiplication'' leads to a significant improvement in EER on all benchmarks compared to using ``concatenation.'' This indicates the superiority of the multiplicative operation, as it promotes interaction between low-level and high-level features during training and contributes to enhanced performance \cite{RN37}. For model training, the original approach involves training each set of corresponding features with individual co-tuplet losses. In contrast, an alternative approach is employed where the features are concatenated and trained with only one loss. The results in Table~\ref{tab:5} reveal that training ``with individual losses'' outperforms training ``with only one loss'' across all benchmarks. Furthermore, using an alternative operation for model training significantly impacts performance more than using a different operation for the feature fusion strategy. This result suggests that training ``with individual losses'' enables the model to learn more discriminative representations and improves overall performance.

\begin{table*}[!t]\centering
	\caption{Signature verification performance without each module/branch in the proposed framework (EER in \%). Values in parentheses indicate the performance degradation (EER increase in \%) compared to using each module/branch}\label{tab:4}
	\begin{tabular}[t]{lK{1.6cm}K{1.6cm}K{1.6cm}K{1.6cm}}
		\toprule
		\multirow{2}*{without Module/Branch} & \multirow{2}*{CEDAR} & BHSig- & BHSig- & \multirow{2}*{HanSig}\\
		& & Bengali & Hindi & \\ 
		\midrule
		MFF module & 6.74 (3.23) & 7.29 (1.17) & 9.84 (3.16) & 11.02 (1.09)\\
		GRCA module & 4.78 (1.27) & 6.59 (0.47) & 6.96 (0.28) & 10.48 (0.55)\\
		Global branch & 6.70 (3.19) & 7.34 (1.22) & 7.19 (0.51) & 10.70 (0.77)\\
		Regional branch & 5.14 (1.63) & 8.26 (2.14) & 10.85 (4.17) & 10.76 (0.83)\\
		\bottomrule
	\end{tabular}
\end{table*}

\begin{table*}[!t]\centering
	\caption{Performance comparison between different operations in our approach (EER in \%)}\label{tab:5}
	\begin{tabular}[t]{llcccc}
		\toprule
		& \multirow{2}*{Operation} & \multirow{2}*{CEDAR} & BHSig- & BHSig- & \multirow{2}*{HanSig}\\
		& & & Bengali & Hindi & \\
		\midrule
		\multirow{2}*{Feature fusion strategy} & Concatenation & 7.83 & 7.49 & 8.81 & 10.12\\
		& Multiplication (Ours) & 3.51 & 6.12 & 6.68 & 9.93\\
		\midrule
		\multirow{2}*{Model training} & With only one loss & 8.55 & 7.93 & 9.31 & 10.47\\
		& With individual losses (Ours) & 3.51 & 6.12 & 6.68 & 9.93\\
		\bottomrule
	\end{tabular}
\end{table*}

\begin{figure*}[htb]
	\centering
	\subfloat[CEDAR]{\includegraphics[width=0.45\linewidth]{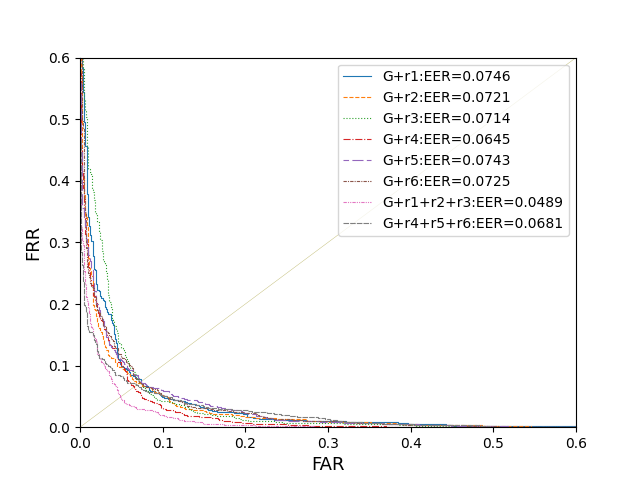}}
	\hfill
	\subfloat[BHSig-Bengali]{\includegraphics[width=0.45\linewidth]{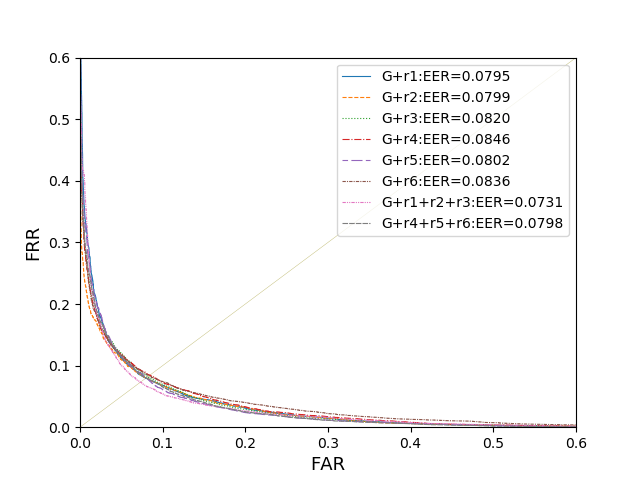}} \\
	\subfloat[BHSig-Hindi]{\includegraphics[width=0.45\linewidth]{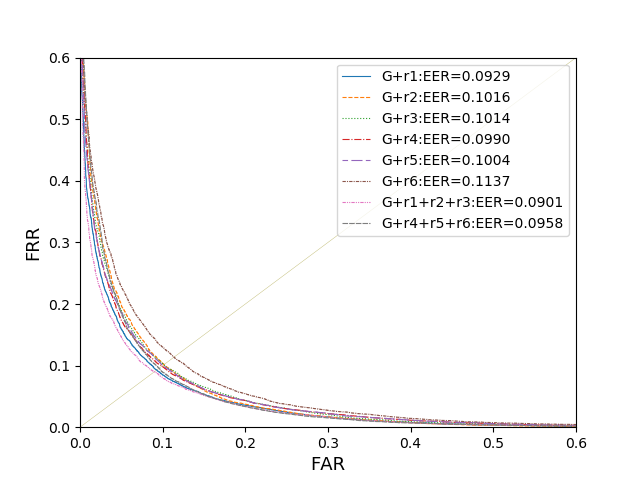}}
	\hfill
	\subfloat[HanSig]{\includegraphics[width=0.45\linewidth]{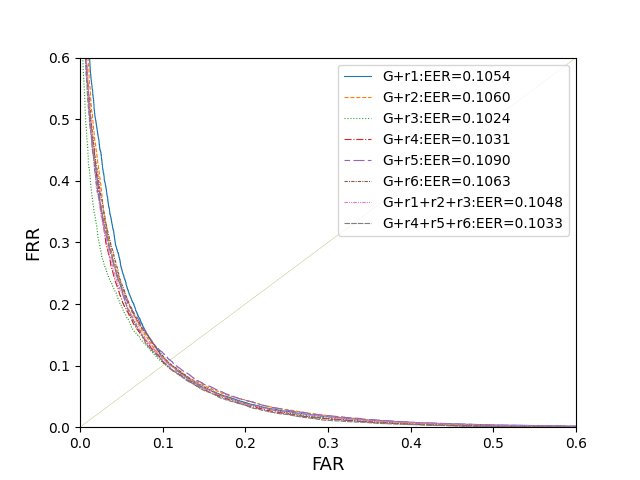}}
	\\[5pt]
	\caption{ROC curves of global and regional feature learning using different individual regions and combinations of regions. Here, G and r$_i$ denote global feature learning and regional feature learning from $F_{r_i}$, respectively}
	\label{fig7}
\end{figure*}

\bigskip\noindent
\textbf{Effect of different regions} We evaluate the effect of using different individual regions and combinations of regions for multiscale feature learning. In this ablation experiment, we use various regions within the regional branch to complement the global features and assess performance across four datasets. Additionally, we use a combination of vertical regions ($F_{r_1}$, $F_{r_2}$, and $F_{r_3}$), as well as a combination of horizontal regions ($F_{r_4}$, $F_{r_5}$, and $F_{r_6}$), to complement the global features and compare the performance. The resulting ROC curves are shown in Fig. \ref{fig7}. We observe from the results that using $F_{r_1}$ or $F_{r_4}$ as complements to the global features generally achieves better performance across most datasets. Furthermore, incorporating $F_{r_2}$ or $F_{r_3}$ also yields favorable outcomes. This may suggest that signatures from the same writer show less variability in these regions, or that forged signatures differ more distinctly from genuine ones in these areas, thereby offering more discriminative information. We also observe that combining vertical or horizontal regions generally achieves better performance than using a single vertical or horizontal region alone. Additionally, combinations of vertical regions outperform combinations of horizontal regions on most datasets.

\subsection{Visualization analysis}
\subsubsection{Comparison between extracted features}
\begin{figure*}[!t]
	\begin{minipage}[t]{0.29\linewidth}
		\centerline{\includegraphics[height=4.4cm]{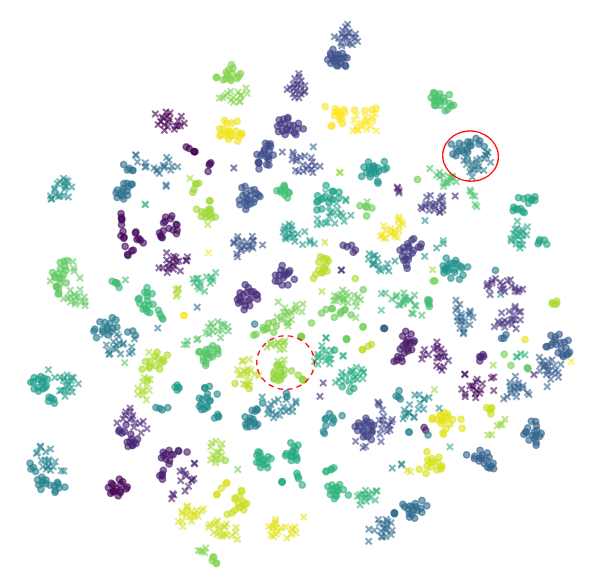}}
		\centerline{\footnotesize (a) Simple baseline (pretrained VGG-16)}
	\end{minipage}
	\hfill
	\begin{minipage}[t]{0.4\linewidth}
		\centerline{\includegraphics[height=4.3cm]{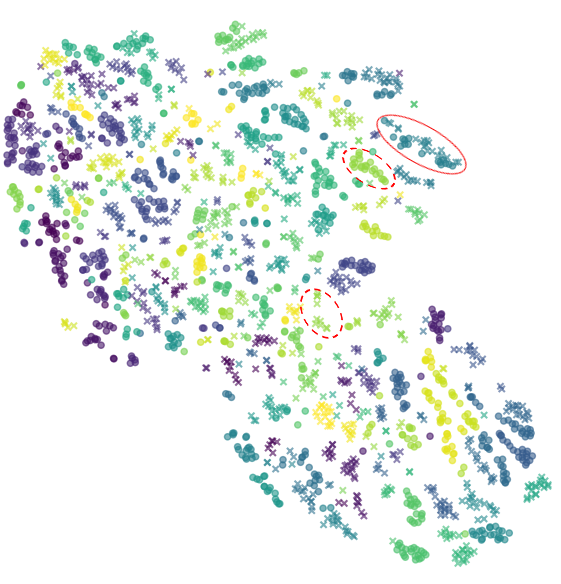}}
		\centerline{\footnotesize (b) MS-SigNet+triplet loss}
	\end{minipage}
	\hfill
	\begin{minipage}[t]{0.29\linewidth}
		\centerline{\includegraphics[height=4.4cm]{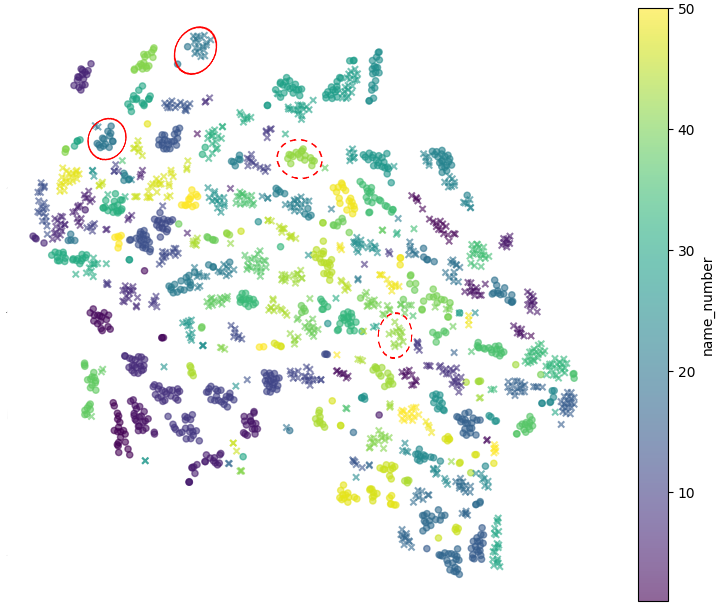}}
		\centerline{\footnotesize (c) MS-SigNet+co-tuplet loss}
	\end{minipage}
	\\[3pt]
	\caption{2D projections of the extracted features of the random 50 names (each name has 20 genuine and 20 skilled-forged signature images) from the HanSig test set using t-SNE \cite{RN46}. Each marker represents a signature sample: ``o'' represents the genuine signatures, and ``x'' represents the forged signatures. The signature samples belonging to different names are displayed in different colors. The red solid circle and dashed circle indicate the samples of the two names} 
	\label{fig8}
\end{figure*}

We compare the 2D projections of extracted features from the simple baseline (pretrained VGG-16), MS-SigNet with triplet loss, and MS-SigNet with co-tuplet loss using the t-distributed stochastic neighbor embedding (t-SNE) algorithm \cite{RN46}. To ensure clarity, we use signatures from a randomly selected subset of 50 names out of the 90 names in the HanSig test set. Fig. \ref{fig8} (a) displays the feature space projection of the simple baseline. It shows that genuine and corresponding forged signatures of each name are clustered together. For instance, the red solid circle highlights a cluster from a single name where genuine signatures are visually inseparable from forged signatures represented by ``o'' and ``x'' marks, respectively. A similar pattern is observed for another name within the red dashed circle. This result suggests that the simple baseline can differentiate between signatures of different names, distinguishing genuine signatures from random forgeries rather than effectively distinguishing between genuine signatures and skilled forgeries.

Fig. \ref{fig8} (b) presents the feature space of MS-SigNet trained with the triplet loss. It demonstrates a better separation between genuine signatures and skilled forgeries compared to the simple baseline. However, some genuine signatures and their corresponding forgeries remain clustered together. For example, the two red dashed circles indicate separate genuine and forged signatures of a name, while the red solid circle shows that genuine and forged signatures are visually indistinguishable. Additionally, the triplet loss fails to pull genuine signatures of each name closer together. Fig. \ref{fig8} (c) illustrates the feature space of MS-SigNet trained with the co-tuplet loss. It shows visually separable clusters of genuine signatures and skilled forgeries for each name. The two red solid circles highlight separate clusters of genuine and forged signatures for one name, while the two red dashed circles indicate a similar pattern for another name. These experimental results demonstrate the promising generalization ability of our proposed MS-SigNet with the co-tuplet loss to unseen data.

\subsubsection{Comparison between global and regional branches}
\begin{figure}[!t]
	\begin{minipage}{0.49\linewidth}
		\centerline{\includegraphics[height=5.5cm]{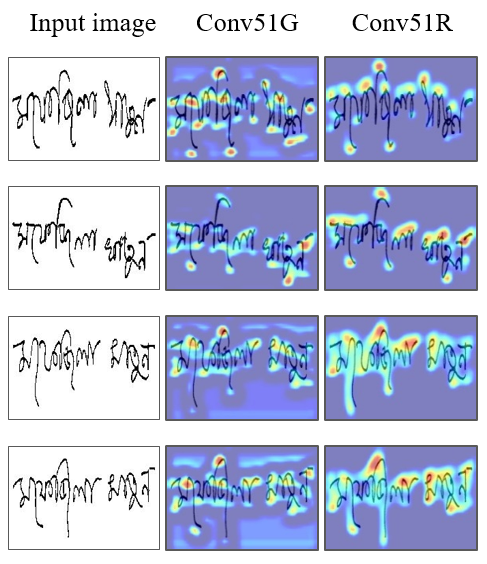}}
		\centerline{(a) BHSig-Bengali}
	\end{minipage}
	\hfill
	\begin{minipage}{0.49\linewidth}
		\centerline{\includegraphics[height=5.5cm]{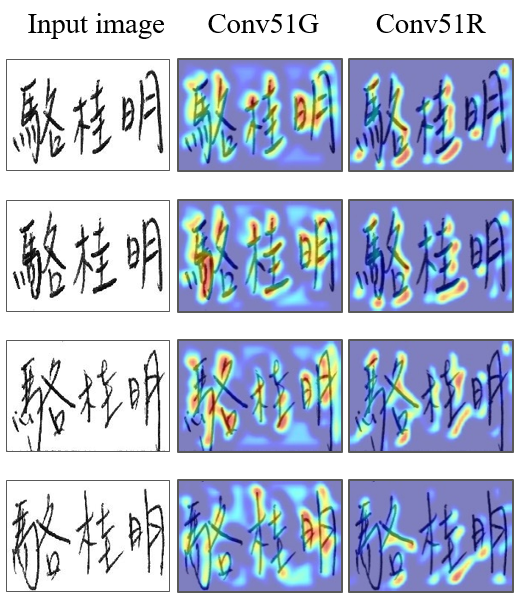}}
		\centerline{(b) HanSig}
	\end{minipage}
	\\[3pt]
	\caption{Visualizing the convolutional layer of the global branch and the regional branch using Grad-CAM \cite{RN47, RN48}. The images in the first column are the input images. The top two and bottom two images are genuine signatures and the corresponding skilled forgeries, respectively, from the test set of BHSig-Bengali and HanSig. The visualization results generated from Conv51G and Conv51R are shown in the second and third columns} 
	\label{fig9}
\end{figure}

To highlight the benefits of multiscale feature learning, we provide visual explanations of the convolutional layers in both the global and regional branches. We generate heat maps using Grad-CAM \cite{RN47, RN48} for two genuine signature images and their corresponding forgeries from the test sets of BHSig-Bengali and HanSig. Fig. \ref{fig9} displays the results obtained from Conv51G and Conv51R (the first convolutional layers of the global and regional branches).

Conv51G exhibits attention to various areas of the entire image, capturing overall signature information such as outlines and stroke configuration. In comparison, Conv51R focuses more on signature strokes and highlights local details in specific signature regions. For instance, it emphasizes sharp curves in the upper region of the BHSig-Bengali images and slanting lines in the lower region of the HanSig images. Moreover, Conv51R emphasizes both small details (e.g., the end of a vertical line in genuine signatures of BHSig-Bengali) and larger detailed parts (e.g., the entire vertical line in forged signatures of BHSig-Bengali), enabling the model to learn fine-grained differences between signatures. These visualization results indicate that the global and regional branches have distinct focuses and capture different yet complementary signature information. Integrating information obtained from multiple spatial scales allows for the generation of more discriminative features in signature verification.

\subsection{Comparison with state-of-the-art methods}
\subsubsection{Performance comparison}
We present a comparison between our proposed method and several state-of-the-art methods on the three public datasets. For our newly created HanSig, we provide results of several baseline methods for comparison under the same experimental settings. The results are summarized in Tables~\ref{tab:6} and \ref{tab:7}, respectively. Note that previous works might report different metrics for their methods. To ensure comparability across different studies, we focus on EER and AUC. In cases where previous works only report Average Error Rate (AER), the average of FRR and FAR, which is considered to be comparable to EER \cite{RN2}. We also provide additional information about each compared method. Since WD-based and WI-based systems employ different training and evaluation schemes, we include WD methods only as a reference. It is important to mention that some previous works, as noted by \cite{RN2}, did not perform noise removal on the image background during data preprocessing. However, as shown in Fig. \ref{fig6}, the image backgrounds of genuine and forged signatures in CEDAR exhibit significant differences. Therefore, we exclude works that reported a 0\% error rate without background removal.

Table~\ref{tab:6} demonstrates that our MS-SigNet with the co-tuplet loss achieves superior performance compared to other methods on CEDAR. Our method outperforms OC-SVM \cite{RN49}, graph-based CNN \cite{RN11}, P2S metric \cite{RN30}, Siamese network \cite{RN10}, and MSDN \cite{RN2}, which also employ metric learning for signature verification. Notably, our approach surpasses MSDN \cite{RN2}, which solely utilizes local regions from input segmentation for feature learning. This result highlights the efficacy of our proposed multiscale feature learning.

Table~\ref{tab:6} highlights the competitive result of our proposed MS-SigNet with the co-tuplet loss on BHSig-Bengali. Our method surpasses SigNet \cite{RN9}, SURDS \cite{RN40}, and DeepHSV \cite{RN43} by 7.77, 6.54, and 5.8 percentage points in terms of EER, respectively. These three methods employ typical metric learning losses mentioned in Section \ref{method-subsub1} for signature verification. IDN \cite{RN25} achieves an AER of 4.68\%, which is better than our method on BHSig-Bengali. However, our proposed method outperforms IDN on the CEDAR and BHSig-Hindi datasets. 

The proposed MS-SigNet and co-tuplet loss achieve substantial performance improvement on BHSig-Hindi. Among the WI methods, our method achieves an EER of 6.68\%, which is significantly lower than the EERs of 2C2S \cite{RN52} (9.32\%), SigNet \cite{RN9} (15.36\%), SURDS \cite{RN40} (10.50\%), and DeepHSV \cite{RN43} (13.34\%), and the AER of IDN \cite{RN25} (6.96\%). This comparison with other state-of-the-art methods showcases the competitiveness and effectiveness of our proposed approach.

Since HanSig is a newly-created dataset, we compare the proposed method against three baselines on this dataset. The first baseline is SigNet \cite{RN9}, which utilizes a Siamese network architecture for offline signature verification. We use the provided source code from the author and ensure that the data subsets used for SigNet are consistent with our method. The second and third baselines are VGG-16 \cite{RN45} and ResNet-18 \cite{RN50} pretrained on the ImageNet dataset, respectively. As depicted in Table~\ref{tab:7}, our proposed method demonstrates superior performance on HanSig when compared to the three baseline methods.

\begin{table*}[!t] \centering
	\caption{Comparison with existing methods on the CEDAR, BHSig-Bengali, and BHSig-Hindi datasets (evaluation metrics in \%)}\label{tab:6}
	\renewcommand\arraystretch{1.2}
	\begin{tabular}[t]{lK{0.8cm}K{0.8cm}K{0.8cm}K{0.8cm}K{0.8cm}K{0.8cm}K{0.8cm}K{0.8cm}K{0.8cm}}
		\toprule
		\multirow{2}{*}{Method} & \multirow{2}{*}{Type} & \multirow{2}{*}{\#Ref} & \multirow{2}{*}{Metric} & 
		\multicolumn{2}{c}{CEDAR} & \multicolumn{2}{c}{BHSig-Bengali} & \multicolumn{2}{c}{BHSig-Hindi}\\
		\cmidrule(lr){5-6} \cmidrule(lr){7-8} \cmidrule(lr){9-10}
		& & & & EER & AUC & EER & AUC & EER & AUC\\
		\midrule
		Genetic algorithm \cite{RN7} & WD & 12 & N & 4.67* & - & - & - & - & -\\
		SigNet-F \cite{RN22} & WD & 12 & N & 4.63 & - & - & - & - & -\\
		Texture features \cite{RN19} & WD & 12 & N & - & - & 33.82 & - & 24.47 & -\\
		OC-SVM \cite{RN49} & WD & 12 & Y & 5.60** & - & - & - & - & -\\
		Micro deformations \cite{RN39} & WD & 8 & N & - & - & 8.21 & - & 9.01 & -\\
		Duplication model \cite{RN51} & WD & 2 & N & - & - & 10.67 & 95.30 & 11.88 & 94.15\\
		Graph-based CNN \cite{RN11} & WI & 10 & Y & 12.27 & - & - & - & - & - \\
		P2S metric \cite{RN30} & WI & 5 & Y & 9.29 & - & - & - & - & -\\
		Morphology \cite{RN41} & WI & 1 & N & 11.59 & - & - & - & - & -\\ 
		Surroundedness \cite{RN42} & WI & 1 & N & 8.33* & - & - & - & - & -\\
		IDN \cite{RN25} & WI & 1 & N & 3.62 & - & {\textbf{4.68}}** & - & 6.96** & -\\
		2C2S \cite{RN52} & WI & 1 & N & - & - & 6.75 & - & 9.32 & -\\   
		SigNet \cite{RN9} & WI & 1 & Y & - & - & 13.89* & - & 15.36* & -\\
		SURDS \cite{RN40} & WI & 1 & Y & - & - & 12.66** & - & 10.50** & -\\
		DeepHSV \cite{RN43} & WI & 1 & Y & - & - & 11.92 & 95.50 & 13.34 & 94.00\\
		Siamese network \cite{RN10} & WI & 1 & Y & 8.50 & - & - & - & - & -\\
		MSDN \cite{RN2} & WI & 1 & Y & 6.74 & - & - & - & - & -\\
		MS-SigNet+co-tuplet loss & WI & 1 & Y & {\textbf{3.51}} & {\textbf{99.47}} & 6.12 & {\textbf{98.64}} & {\textbf{6.68}} & {\textbf{98.28}}\\
		\bottomrule
	\end{tabular}
	\footnotetext{Type refers to the WD or WI approach. \#Ref indicates the number of genuine signatures as the references for the questioned signature to be compared with. Metric refers to whether metric learning is used for model training, with ``Y'' indicating yes and ``N'' indicating no. ``*'' refers to AER when FRR equals FAR; hence, it is the same as EER. ``**'' refers to AER, but FRR and FAR are of unequal value.}
	\footnotetext{The best results are marked in boldface.}
\end{table*}

\begin{table}[!t]\centering
	\caption{Comparison with baseline methods on the HanSig dataset (evaluation metrics in \%)}\label{tab:7}
	\renewcommand\arraystretch{1.2}
	\begin{tabular}[t]{lcccc}
		\toprule
		Method & FRR & FAR & EER & AUC\\
		\midrule
		SigNet \cite{RN9} & 32.43 & 19.66 & 26.31 & 80.94\\
		Pretrained VGG-16 \cite{RN45} & 19.18 & 36.92 & 28.86 & 78.34\\
		Pretrained ResNet-18 \cite{RN50} & 27.57 & 37.05 & 32.42 & 73.77\\
		MS-SigNet+co-tuplet loss & {\textbf{7.69}} & {\textbf{11.85}} & {\textbf{9.93}} & {\textbf{96.38}}\\
		\bottomrule
	\end{tabular}
	\footnotetext{The best results are marked in boldface.}
\end{table}

\subsubsection{Model complexity comparison}
Table~\ref{tab:8} compares the model complexity of our method with other methods in terms of the number of trainable parameters. The backbone of our offline signature verification system is a modified version of the SigNet-F \cite{RN22} structure. It includes only five convolutional layers and one FC layer, making it more parameter-efficient compared to very deep networks such as VGGNet \cite{RN45} and Inception \cite{RN53}. Consequently, the proposed MS-SigNet is more lightweight than systems \cite{RN24, RN52, RN43, RN40, RN11} that either use very deep networks as backbones or adopt them directly. MS-SigNet can be trained on small signature datasets to achieve highly competitive performance, even without pretraining on large datasets.

\begin{table}[h]\centering
	\caption{Model complexity comparison between different methods in terms of the number of trainable parameters}\label{tab:8}%
	\begin{tabular}{@{}lll@{}}
		\toprule
		Method & Backbone & \#Params (M) \\
		\midrule
		GWBTA+CNN\cite{RN24} & VGG-16 \cite{RN45} & 151.04/138.36 \\
		2C2S \cite{RN52} & Swin Transformer \cite{RN54} & 47.40/29.00 \\
		DeepHSV \cite{RN43} & Inception-v3 \cite{RN53} & 21.81/21.81 \\
		SURDS \cite{RN40} & ResNet-18 \cite{RN50} & 22.88/11.69 \\
		Graph-based CNN \cite{RN11} & DenseNet-121 \cite{RN55} & 7.98/7.98 \\
		Proposed MS-SigNet & Modified SigNet-F \cite{RN22} & 7.96/3.99 \\
		\bottomrule
	\end{tabular}
	\footnotetext{\#Params denotes the number of trainable parameters. This column's first and second values are the number of trainable parameters of a method and its backbone.}
	\footnotetext{The number of parameters in the SURDS is calculated based on the encoder and projector components.}
\end{table}

\section{Conclusion}
In this study, we propose a multiscale feature learning network and a new metric learning loss to build an automatic handwritten signature verification system. The proposed MultiScale Signature feature learning Network (MS-SigNet) captures complementary signature information from multiple spatial scales. It can integrate the information to generate discriminative features for static signature verification. The multilevel feature fusion and global-regional channel attention (GRCA) modules designed for the two-branch structure provide further performance gains. To enhance the discriminative capability of our verification system, we propose the co-tuplet loss, a novel metric learning loss function. Experimental results demonstrate that our MS-SigNet with the co-tuplet loss surpasses the state-of-the-art methods on various benchmark datasets, showcasing its effectiveness in signature verification across different languages. While our results are promising, further improvements could be made by developing alternative methods to integrate multiple signature information.

\bibliographystyle{IEEEtran}
\bibliography{bibtexRef-AI}


\end{document}